\documentclass[twoside,leqno,twocolumn]{article}
\usepackage[letterpaper]{geometry}
\pdfoutput=1 
\usepackage{ltexpprt}
\usepackage{hyperref}
\usepackage{graphicx}
\usepackage{amssymb}
\usepackage{amsmath}
\usepackage{bm}
\usepackage{biblatex}
\usepackage{algorithm}
\usepackage{algorithmic}
\usepackage{soul}
\usepackage{subcaption}
\usepackage{bbm}
\usepackage{array,amsfonts}
\usepackage{multicol}
\usepackage{subfiles}

\begin{document}

\title{\Large Personalized Federated Learning with Contextual Modulation and Meta-Learning}
\author{Anna Vettoruzzo\thanks{Center for Applied Intelligent Systems Research (CAISR), Halmstad University, Halmstad, Sweden.}\\ \small anna.vettoruzzo@hh.se 
\and Mohamed-Rafik Bouguelia\footnotemark[1] \\ \small mohamed-rafik.bouguelia@hh.se 
\and Thorsteinn Rögnvaldsson\footnotemark[1] \\ \small thorsteinn.rognvaldsson@hh.se }

\date{}

\maketitle

% Default Copyright Statement
\fancyfoot[R]{\scriptsize{Copyright \textcopyright\ 2024 by SIAM\\
Unauthorized reproduction of this article is prohibited}}

\begin{abstract} \small\baselineskip=9pt
Federated learning has emerged as a promising approach for training machine learning models on decentralized data sources while preserving data privacy. However, challenges such as communication bottlenecks, heterogeneity of client devices, and non-i.i.d. data distribution pose significant obstacles to achieving optimal model performance. We propose a novel framework that combines federated learning with meta-learning techniques to enhance both efficiency and generalization capabilities. Our approach introduces a federated modulator that learns contextual information from data batches and uses this knowledge to generate modulation parameters. These parameters dynamically adjust the activations of a base model, which operates using a MAML-based approach for model personalization. Experimental results across diverse datasets highlight the improvements in convergence speed and model performance compared to existing federated learning approaches. These findings highlight the potential of incorporating contextual information and meta-learning techniques into federated learning, paving the way for advancements in distributed machine learning paradigms.
\end{abstract}


%===================================================================
\section{Introduction}
Machine learning and deep learning have revolutionized various domains by enabling models to learn from vast amounts of centralized data. However, the traditional approach of collecting data into a central server raises concerns regarding data privacy and security. To address these issues, federated learning (FL) \cite{mcmahan2017communication} has emerged as a promising framework that enables collaborative model training while keeping the data decentralized on individual devices. %FL distributes the training process across multiple clients or devices, allowing them to learn from their local data while contributing to the improvement of a shared global model at the server level. 
This decentralized approach has gained significant attention in domains such as healthcare, Internet of Things (IoT), and mobile applications, where data cannot be easily transferred to a central server due to privacy regulations, connectivity limitations, or security risks.  
However, in the context of FL, data distributions across clients are often not ``\emph{independent and identically distributed}" (non-i.i.d.) since different devices generate and collect data separately. This poses a challenge for FL methods, such as FedAvg \cite{mcmahan2017communication}, that were primarily designed to handle i.i.d. data. 
Personalized federated learning (PFL) \cite{fallah2020personalized} takes a step further by customizing models to the needs and preferences of individual clients, an important aspect in domains such as recommendation systems and user behavior analysis.
%By capturing the data distribution and local objectives of each client, PFL enables personalized predictions and enhances user experiences in applications where personalization is needed, such as recommendation systems and user behavior analysis. 
One way to accomplish this is by integrating meta-learning techniques into the FL framework. For example, Per-FedAvg \cite{fallah2020personalized} finds an initialization of a global model that performs well after each client's update with respect to its own data and loss function, potentially by performing a few steps of gradient descent. This approach accounts for the heterogeneity of data distributions while benefiting from collaborative training. 
However, when there are substantial differences between client distributions, relying on a single global model may be insufficient. %Consequently, achieving global model convergence and generalization across clients becomes a challenging task, potentially resulting in degraded performance due to the negative knowledge transfer. 
Some approaches have attempted to tackle this issue by learning multiple global models that simultaneously fit different data distributions \cite{duan2021flexible, ghosh2020efficient,  mansour2020three, sattler2020clustered, yang2023personalized}, but this can impose a high computational cost at the server level, as storing multiple models typically requires significant memory resources. Moreover, selecting the most suitable model for each client based on the limited locally available data can be challenging.

In this paper, we propose \emph{CAFeMe}, a \textbf{c}ontext-\textbf{a}ware \textbf{fe}derated learning solution that leverages \textbf{me}ta-learning to facilitate personalization to each client.
Our approach is based on a global model comprising a federated modulator and a base model. The federated modulator network captures contextual information from the data of each client and generates modulation parameters that dynamically adjust the activations of the base model.
In each communication round, the global model is distributed to the clients, 
which personalize it with their locally available data. The modulation parameters generated by the federated modulator serve as additional parameters within the base network, allowing for the personalization of the model to the unique characteristics of each client's data distribution. This allows our proposed approach to deal with high heterogeneity in the clients' data while maintaining a single model at the server level. %After personalization, the clients send their updates back to the server, where the aggregated updates from different devices are used to update the global model. 

%Through extensive experiments on various datasets, 
We demonstrate the effectiveness of \emph{CAFeMe} in improving model convergence, generalization, and personalized prediction on various datasets within the PFL framework. The results highlight the potential of leveraging only relevant information for personalization to each client. This also results in a fast learning process that is particularly advantageous in time-sensitive applications. Furthermore, we showcase the robustness of our framework under the concept shift scenario, where the same class-label may be related to different concepts across clients. %Our results highlight the versatility and applicability of \emph{CAFeMe} in real-world scenarios, emphasizing its value in achieving enhanced performance and adaptability in personalized federated learning settings.
 The code is available online at \url{https://github.com/annaVettoruzzo/CAFeMe}. 

%===================================================================
\section{Related works}
%Due to growing privacy concerns and data regulations, federated learning (FL) has emerged as a powerful approach that enables multiple parties to collaboratively train a model without sharing their local data. 
The pioneering federated learning (FL) algorithm, FedAvg \cite{mcmahan2017communication}, %combines local stochastic gradient descent on each client with model averaging on a central server. This approach 
performs well in scenarios where the local data across clients is i.i.d. However, in scenarios with heterogeneous (non-i.i.d.) data distributions, several techniques have been proposed to improve local learning. These include regularization methods \cite{acarfederated,  li2021model, li2020federated}, local bias correction techniques \cite{dandi2022implicit, karimireddy2020scaffold, murata2021bias}, data sharing mechanisms \cite{zhao2018federated}, data augmentation methods \cite{yoonfedmix}, and selective clients aggregation \cite{tang2022fedcor, wu2022node}.
These methods primarily focus on improving the performance of the global model without considering better customization for each client. 
An alternative approach is personalized federated learning (PFL), which aims to develop customized models for individual clients while leveraging the collaborative nature of FL. 
Popular PFL methods include L2GD \cite{hanzely2020federated}, which combines local and global models, FedRep \cite{collins2021exploiting} that decouples the parameters of the feature extractor and the classifier to learn unique local heads for each client, methods that investigate client-specific model aggregations \cite{huang2021personalized, zhangpersonalized}, as well as multi-task learning methods such as pFedMe \cite{t2020personalized}, Ditto \cite{li2021ditto}, and FedPAC \cite{fedpac2023}. Recently, some frameworks propose to incorporate meta-learning into the PFL framework \cite{vettoruzzo2023advances}. One notable approach is Per-FedAvg \cite{fallah2020personalized}, which proposes a decentralized version of model-agnostic meta-learning (MAML) \cite{finn2017model} for the FL setting. MAML is a meta-learning algorithm that learns models that can efficiently be adapted, through an optimization procedure, to new tasks. In the context of PFL, \emph{personalization to a client} aligns with \emph{adaptation to a task} \cite{jiang2019improving}. 
Other papers \cite{chen2018federated, jiang2019improving, lifair} explore the combination of MAML-type methods with FL architectures to improve personalization for each client's local dataset. Additionally, the authors of \cite{khodak2019adaptive} propose ARUBA, a meta-learning algorithm inspired by online convex optimization, which enhances the performance of FedAvg.

While the effectiveness of these approaches in PFL has been demonstrated, concerns remain regarding their ability to handle highly heterogeneous data distributions among clients.  
A special type of PFL approach that tries to address this concern is clustered or group-based FL \cite{duan2021flexible, ghosh2020efficient, mansour2020three, sattler2020clustered, yang2023personalized}, where multiple (i.e., $K$) group-based global models are learned after partitioning clients into $K$ groups. These methods can either perform clustering at the server level, \cite{sattler2020clustered, ghosh2019robust, yang2023personalized}, or let the clients compute the cluster identity based on their local empirical loss \cite{ghosh2020efficient}. However, partitioning clients into groups presents challenges due to the ambiguity in the clustering structure, and the memory costs of storing $K$ models at the server level. 

Our contribution consists of improving the performance and scalability of the PFL framework, even in scenarios where clients exhibit substantial differences in the data distribution. Our proposed method incorporates context information extracted from each client's local data to drive the personalized model towards a better fit of the client-specific data distribution, while ensuring that the personalized base model does not deviate too far from the global version at the server level. This approach ensures a balance between local adaptation and global consistency, facilitating effective model personalization for each client.

\section{Background and Notation}
This section provides a formal description of the PFL problem and its connection with the MAML algorithm \cite{finn2017model}, specifically designed for meta-learning problems.
In PFL, we consider a distributed learning setting where a central server communicates with $C$ clients. 
Let $\mathbb{C}=\{1,\cdots, C\}$ denote the set of clients participating in the training process. Each client $i$ has its own data distribution $P_{\mathcal{X}\mathcal{Y}}^{(i)}$ on $\mathcal{X} \times \mathcal{Y}$, where $\mathcal{X}$ represents the input space and $\mathcal{Y}$ denotes the label space. Each client $i$ has access to $n_i$ data points, represented as $\mathcal{D}_i = \{x_k, y_k\}$, which are samples drawn from $P_{\mathcal{X}\mathcal{Y}}^{(i)}$. 
In this context, it is assumed that the probability distributions $P_{\mathcal{X}\mathcal{Y}}^{(i)}$ and $P_{\mathcal{X}\mathcal{Y}}^{(j)}$ differ for any pair of clients $i$ and $j$, which is typically the case in PFL. This is referred to as the non-i.i.d setting. The differences in probability distributions can arise from variations in $P_{\mathcal{X}}$, in $P_{\mathcal{Y}}$ or in the conditional $P_{\mathcal{Y}|\mathcal{X}}$, resulting in covariate shift, label shift, or concept shift, respectively. While previous works have primarily focused on the first two, our proposed solution is designed to effectively handle the concept shift scenario as well. 

The main goal of PFL is to learn a global model $f_{\omega}$ that performs well after personalization to each client's local dataset. This is achieved through an update scheme that proceeds in rounds of communication. In each communication round of the training process, the global model is initialized with $\omega$ and sent to the clients. Each client $i \in \mathbb{C}$ then personalizes the global model using its local data $\mathcal{D}_i$ to derive personalized model parameters $\omega'_i$. Typically, this personalization process involves performing a few steps of gradient descent on the client's local dataset. Formally, let $\mathcal{L}_{\mathcal{D}_i}(f_{\omega'_i})$ be the empirical risk of the local model $f_{\omega'_i}$ on the local dataset $\mathcal{D}_i$. The objective of PFL can be described as follows:
\begin{equation}
    \min_{\bm{W}} \biggl\{ F(\bm{W}) := \frac{1}{C} \sum_{i=1}^C \mathcal{L}_{\mathcal{D}_i}(f_{\omega'_i}) \biggr\}
\end{equation}
where $\bm{W}$ denotes the collection of all local models, i.e., $\bm{W} = \{\omega'_i\}_{i=1}^{C}$.
However, in the non-i.i.d setting, it is challenging to find a global model that performs well after personalization to heterogeneous clients. One way to address this challenge is to incorporate meta-learning into the PFL setting. 

Meta-learning approaches aim to learn models that can be efficiently adapted to new tasks by leveraging the knowledge acquired from previous tasks. Optimization-based meta-learning methods, such as MAML \cite{finn2017model}, use a bi-level optimization process to learn an initial set of parameters $\omega$ that can be effectively adapted with gradient descent on new tasks. During the meta-training phase, a set of training tasks $\{\mathcal{T}_i\}_{i=1}^{T}$ is used. Each task $\mathcal{T}_i$ corresponds to data generating distributions $\mathcal{T}_i \triangleq \{ P_{\mathcal{X}}^{(i)}, P_{\mathcal{Y}|\mathcal{X}}^{(i)}\}$. The data sampled from each task is divided into a \emph{support set} $\mathcal{D}_i^{(sp)}$ containing $K$ training examples, and a \emph{query set} $\mathcal{D}_i^{(qr)}$ used for evaluating the model's performance.
MAML meta-trains a neural network model $f_{\omega}$ parameterized by $\omega$ using a two-stage procedure consisting of an inner and an outer loop. In the inner loop, the initial parameters $\omega$ are adapted to each training task $\mathcal{T}_i$ by taking a few gradient descents steps on the support set $\mathcal{D}_i^{(sp)} \triangleq \{x_k, y_k\}_{k=1}^K$. This process results in task-specific parameters $\omega'_i$, as illustrated in Eq. \ref{eqn:adapt_maml} when a single gradient step is used:
\begin{equation}\label{eqn:adapt_maml}
\omega'_i = \omega - \alpha \nabla_\omega\mathcal{L}_{\mathcal{D}_i^{(sp)}}(f_{\omega}).
\end{equation}
The initial parameters $\omega$ are then optimized in the outer loop by minimizing the loss achieved by the task-specific parameters $\omega'_i$ on the query set $\mathcal{D}_i^{(qr)}$ of the same task $\mathcal{T}_i$. This is shown in Eq. \ref{eqn:update_maml}:
\begin{equation} \label{eqn:update_maml}
\omega \leftarrow \omega - \beta \nabla_{\omega} \sum_{\mathcal{T}_i \sim \textnormal{P}(\mathcal{T})} \mathcal{L}_{\mathcal{D}_i^{(qr)}}(f_{\omega'_i}).
\end{equation}
The result is a model initialization $\omega$ that performs well when updated with a few training examples $\mathcal{D}_{new}^{(sp)} \triangleq \{x_k, y_k\}_{k=1}^K$ from a new task $\mathcal{T}_{new}$. The goal is to train the model on $\mathcal{D}_{new}^{(sp)}$ while leveraging the previous knowledge acquired during meta-training in order to achieve good generalization performance on new unlabeled test examples from $\mathcal{T}_{new}$.

As suggested in \cite{jiang2019improving, fallah2020personalized}, optimization-based meta-learning algorithms, such as MAML, bear relevance to the PFL setting due to the inherent heterogeneity present both in tasks (for meta-learning) and in clients (for federated learning). While MAML assumes access to only a few labeled data from a new task at test time, PFL does not have this constraint. Consequently, an estimation of the empirical loss (in Equations \ref{eqn:adapt_maml} and \ref{eqn:update_maml}) can be computed using independent batches of data (e.g., $\mathcal{D}_i'$, $\mathcal{D}_i''$, etc.) that belongs to $\mathcal{D}_i$, as suggested in \cite{fallah2020personalized}. 
By incorporating the MAML formulation into PFL, the objective of PFL shifts towards finding a model initialization shared among all clients that performs well once each client updates it based on the loss function computed with their own local data. 

\section{Proposed approach}
\begin{figure}[tbp]
\centering
\includegraphics[width=1.1\columnwidth]{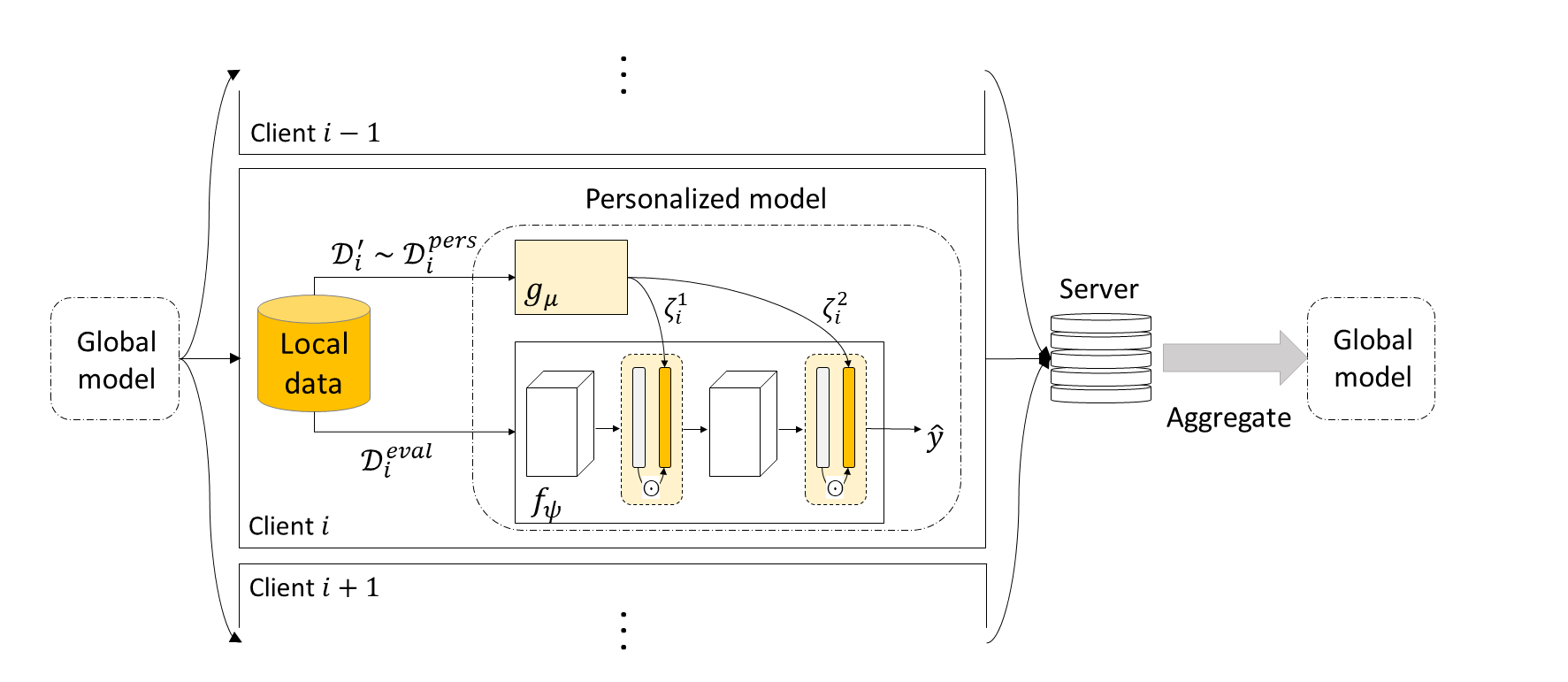} 
\caption{Overall framework of \emph{CAFeMe}. For each client involved in the federation, the federated modulator $g_\mu$ takes in input a batch of local data $\mathcal{D}'_i$, sampled from $\mathcal{D}_i^{pers}$, and predicts the parameters of the modulation layers $\boldsymbol{\zeta}_i$. These parameters are used to modulate the activations of the base network $f_{\psi}$. The modulation operator is denoted by $\odot$.} \label{fig:cafeme}
\end{figure}

In this section, we present our proposed \emph{CAFeMe} approach in detail. The aim is to improve the performance and scalability of the PFL framework by leveraging meta-learning techniques and extracting contextual information from each client's local data. To achieve this, we introduce a federated modulator network $g_{\mu}$ followed by a base network $f_{\psi}$, as shown in Figure \ref{fig:cafeme}.  We define the complete set of model parameters encompassing both the modulator and the base network as $\omega = \{\mu, \psi\}$. Additionally, we incorporate \emph{modulation layers} in the base network to enable efficient personalization to each client's local dataset. These layers aim to condition the base network $f_{\psi}$ on that client's data by employing a feature-wise gating mechanism to the activations of the preceding layers, allowing only the relevant features (or activations) to propagate forward while disregarding the non-relevant ones. The parameters of the modulation layers are predicted by the federated modulator. The latter extracts some context from batches of data available at the client level and transforms it to generate the modulation parameters. The context is designed to be representative of the client's local dataset and invariant to the permutation of data samples, ensuring the predicted modulation parameters are independent of the ordering of the data. We delve into the architecture of the federated modulator in Section \ref{sec:generator} and provide the complete algorithm in Section \ref{sec:algo}.
%The whole set of parameters $\omega = \{\mu, \psi\}$ is broadcast to each client

\subsection{Federated Modulator.} \label{sec:generator}
\begin{figure}[tbp]
\centering
\includegraphics[width=0.45\textwidth]{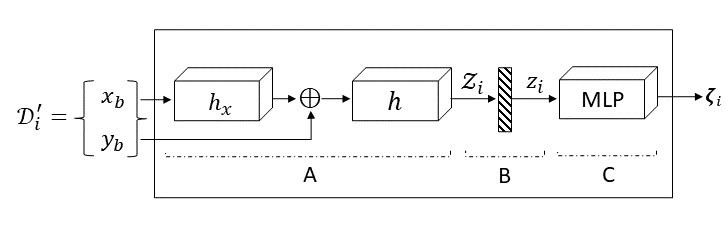} 
\caption{Architecture of the federated modulator. It consists of three parts: a feature extractor (A), an averaging layer (B), and an MLP (C).} \label{fig:modulator}
\end{figure}

As illustrated in Figure \ref{fig:modulator}, the federated modulator network $g_{\mu}$ consists of three parts: a feature extractor (A), an averaging layer (B), and a multilayer perception (MLP) (C). 
The feature extractor (A) takes as input a batch of data $\mathcal{D}_i' = \{x_b, y_b\}_{b=1}^B$, where $B$ is the batch size. Each batch consists of inputs $x_b$ and labels $y_b$ (represented as one-hot-vectors in classification) sampled from a client $i$. It then transforms them into $\mathcal{Z}_i = \{ h( ~ h_x(x_b) ~\oplus~ y_b ~ ) \}_{b=1}^B$, where $h_x(.)$ denotes an input-specific feature extractor, and $\oplus$ denotes the concatenation operator. %In other words, it transforms each sample $\mathbf{x}_b=h_x(x_b)$, it concatenates to it the corresponding label and transforms them as $h( \mathbf{x}_b \oplus y_b )$. 
The averaging layer (B) calculates the mean of the transformed data $\mathcal{Z}_i$ along the $B$ examples, creating a vector representation $z_i$ that characterizes client $i$. This task representation is invariant to permutations of the examples and is expected to capture client-specific information. %, such as an estimation of the client's data distribution.
The final part (C) of the network generates the parameters for the modulation layers in the base network. It is an MLP that takes $z_i$ as input and predicts a set of parameters $\boldsymbol{\zeta}_i = \{ \zeta_i^1, \cdots \zeta_i^L\}$, where $L$ is the number of hidden layers in the base model. Each $\zeta_i^l$, for $l=1, ..., L$, corresponds to the parameters of the modulation layer following the $l^{\textnormal{th}}$ layer in the base model.
These parameters will act as ``scaling" or ``gating" parameters; they adjust the feature importance and diminish the contribution of less informative features for the given client $i$. This is described more formally in the next section.

\subsection{Algorithm.} \label{sec:algo}
The complete algorithm of the \emph{CAFeMe} method is presented in Algorithm \ref{algo1}. At each communication round $t$, with $t=1,\cdots,T$, a subset of $M$ clients is randomly sampled and denoted with $\mathcal{C}^{(t)}$. In line 4 of Algorithm \ref{algo1}, the parameters of the global model $\omega$ are sent to each of the selected clients with the aim of personalizing them to each client's local data. To do so, each client's dataset $\mathcal{D}_i$ is split into two disjoint sets, $\mathcal{D}_i^{pers}$ and $\mathcal{D}_i^{eval}$. The former set is used in line 7 to personalize the global model's parameters $\omega$ to client $i$. This personalization is described in Algorithm \ref{algo_adapt}, where after a first initialization step in line 1, a batch of data is sampled from $\mathcal{D}_i^{pers}$ and input in the federated modulator $g_{\mu}$ to obtain the modulation parameters $\boldsymbol{\zeta}_i$, as described in Section \ref{sec:generator}. Note that if $S > 1$, an independent batch $\mathcal{D}'$ is sampled at each iteration. The modulation parameters $\boldsymbol{\zeta}_i$ are then used to modulate the activations of the base network, resulting in a \emph{modulated model} denoted by $f_{\psi} \vert \boldsymbol{\zeta}_i$ which is more personalized for the client at hand. To better explain the modulated network $f_{\psi} \vert \boldsymbol{\zeta}_i$, let us examine the activations $a(\mathbf{x} ~ \psi^l)$ produced by a particular layer $l$. Using modulation parameters $\zeta_i^l$, the activations can be transformed to produce modulated activations, indicated as $a(\mathbf{x} ~\psi^l) \odot \zeta_i^l$, as exemplified by Equation \ref{eqn:modulation}. This equation outlines a ``sigmoidal gating" mechanism that enables $\zeta_i^l$ to determine which activations are propagated forward to the next layers and which are zeroed out. Note that this approach is not restricted to this type of modulation, but we believe that gating the model's activations is sufficient to specialize the model to specific clients while simultaneously leveraging the knowledge learned from other clients.
\begin{equation} \label{eqn:modulation}
a(\mathbf{x} ~\psi^l) \odot \zeta_i^l = a(\mathbf{x} ~ \psi^l) \otimes \sigma(\zeta_i^l), 
%\mathbf{s}_i^l = a(\mathbf{x} ~\psi^l) \odot \zeta_i^l
\end{equation}
where $\otimes$ denotes an element-wise multiplication and $\sigma$ is the sigmoid function making each value of $\zeta_i^l$ in the range $[0, 1]$. The parameters $\mu$ and $\psi$ are then personalized to the specific client in lines 5 and 6 of Algorithm \ref{algo_adapt}, resulting in client-specific parameters $\mu'$ and $\psi'$, respectively. This personalization is performed in a MAML-based manner, see Equation \ref{eqn:adapt_maml}, but the loss is now computed with a batch of data $\mathcal{D}_i'$ (not necessarily small as in MAML), and with $f_{\psi}|\boldsymbol{\zeta}_i$ that is the base model whose activations have been transformed with the parameters of the modulation layers. This process can be iterated for $S \geq 1$ gradient descent steps, and the personalized parameters $\psi_i'$ and $\boldsymbol{\zeta}_i$ are returned.

In line 8 of Algorithm \ref{algo1}, the initial set of parameters $\omega$ is updated, similarly to Equation \ref{eqn:update_maml}, by minimizing the loss computed on $\mathcal{D}_i^{eval}$ using $f_{\psi'_i} \vert \boldsymbol{\zeta}_i$ (i.e., the evaluation loss achieved by the model after being personalized to client $i$). Here, any optimizer of choice, e.g., Adam, can be used (not necessarily gradient descent). This results in a set of parameters $\omega_i'$ (comprising both $\mu_i'$ and $\psi_i'$) that are personalized to the client's data and, at the same time, generalize well to a held-out dataset sampled from the same distribution. This guarantee good performance when new data from the same client are observed and prevents overfitting to the client's initial data. 
The updated parameters $\omega_i'$ are then sent back to the server in line 9. After receiving the parameters from all the selected clients, the server collects the local models and computes the average to reinitialize $\omega$, following a similar approach to the FedAvg algorithm \cite{mcmahan2017communication}.

\begin{algorithm} [tb]
\caption{\emph{CAFeMe}}\label{algo1}
\textbf{Require} Active clients $M$, step sizes $\alpha, \beta$
\begin{algorithmic}[1]
\STATE Randomly initialize $\omega = \{\mu, \psi\}$
\FOR {$t= 1, \cdots, T$}
    \STATE Choose a subset of clients $\mathcal{C}^{(t)}$ with size $M$
    \STATE Send $\omega$ to all clients in $\mathcal{C}^{(t)}$
    \FORALL {$i \in \mathcal{C}^{(t)}$}
        \STATE Split data $\mathcal{D}_i = \{\mathcal{D}_i^{pers}, \mathcal{D}_i^{eval}\}$ 
        \STATE Personalize the global model: \\$\psi_i', \boldsymbol{\zeta}_i \gets \emph{Personalization}(S, \alpha, \mathcal{D}_i^{pers}, \omega)$    
        \STATE Update $\omega_i' \gets \omega - \beta \nabla_{\omega} \mathcal{L}_{\mathcal{D}_i^{eval}} (f_{\psi_i'} | \boldsymbol{\zeta}_i)$
        \STATE Send $\omega_i'$ back to the server
    \ENDFOR
    \STATE Server updates $\omega = \frac{1}{M} \sum_{i \in \mathcal{C}^{(t)}} \omega_{i}'$
\ENDFOR
\end{algorithmic}
\end{algorithm}

\begin{algorithm} [tb]
\caption{\emph{Personalization}($S, \alpha, \mathcal{D}^{pers}, \omega$)}\label{algo_adapt}
\textbf{Require} Personalization steps $S$, step size $\alpha$, local dataset for personalization $\mathcal{D}_i^{pers}$, initial parameters of the model $\omega = \{\mu, \psi \}$ 
\begin{algorithmic}[1]
\STATE Initialize $\mu' \gets \mu$, $\psi' \gets \psi$
\FOR {$s=1, \cdots, S$}
    \STATE Sample batch of data $\mathcal{D}'$ from $\mathcal{D}^{pers}$
    \STATE Predict modulation parameters $\boldsymbol{\zeta} = g_{\mu'}(\mathcal{D}')$
    \STATE $\mu' = \mu' - \alpha \nabla_{\mu'} \mathcal{L}_{\mathcal{D}'}(f_{\psi'} \vert \boldsymbol{\zeta})$
    \STATE $\psi' = \psi' - \alpha \nabla_{\psi'} \mathcal{L}_{\mathcal{D}'}(f_{\psi'} \vert \boldsymbol{\zeta})$
\ENDFOR
\STATE Predict modulation parameters $\boldsymbol{\zeta} = g_{\mu'}(\mathcal{D}')$
\STATE \textbf{return} $\psi', \boldsymbol{\zeta}$
\end{algorithmic}
\end{algorithm}

During the testing phase, new clients can join the federation, possibly with local data sampled from different data distributions compared to those encountered during training. The objective is to obtain a personalized model for each new client that performs effectively on its local data, potentially requiring a few personalization steps and only a small amount of labeled data.

\section{Experimental setup} \label{subsec:setup}

\subsection{Datasets.} 
We assess the performance of \emph{CAFeMe} on various datasets, including synthetic and realistic non-i.i.d datasets. For the synthetic datasets, we synthesize distributed non-i.i.d datasets by partitioning well-known benchmarks, such as CIFAR-10 \cite{krizhevsky2009learning} and RMNIST (a modified version of the classic MNIST dataset \cite{deng2012mnist}), into smaller subsets specific to each client. Inspired by previous works in the field \cite{mcmahan2017communication, yurochkin2019bayesian, hsu2019measuring, wang2020federated}, we explore two different data partitioning schemes: the \emph{shards} partition and the \emph{dirichlet} partition. 

We also evaluate the performance of the proposed approach on two more realistic datasets, namely FEMNIST \cite{caldas2018leaf} and Meta-Dataset \cite{triantafilloumeta}. While FEMNIST has been extensively used in the FL literature, this study represents the first time Meta-Dataset is employed to evaluate the performance of a PFL algorithm. Meta-Dataset is a large-scale dataset comprising various classification datasets.
%Meta-Dataset is a large-scale dataset comprising various classification datasets such as Mini-Imagenet \cite{ravi2017optimization}, FC100 \cite{oreshkin2018tadam}, Omniglot \cite{lake2011one}, Aircraft \cite{maji2013fine}, CUB Birds \cite{wah2011caltech}, Describable Textures Dataset (DTD) \cite{cimpoi2014describing}, Traffic Signs \cite{stallkamp2012man}, and VGG Flowers \cite{nilsback2008automated}. 
In this setting, we assume clients share the common tasks of classifying images into $N$ different classes, but their local data are sampled from completely different datasets. For this reason, the same label might have different meanings across clients, hence the concept shift scenario. This scenario closely resembles real-world situations where clients that participate in the federation might belong to different enterprises with diverse goals and interests. %More details about the datasets can be found in the supplementary material.

\subsection{Compared methods.}
We compare the results with several state-of-the-art approaches. Specifically, we include FL works such as FedAvg \cite{mcmahan2017communication} with its fine-tuned variant (i.e., FedAvg-FT), and IFCA \cite{ghosh2020efficient}, a framework specifically designed for clustered FL, together with IFCA-FT. Furthermore, we consider Ditto \cite{li2021ditto} a multi-task learning framework tailored for PFL, FedRep \cite{collins2021exploiting} a PFL method characterized by parameter decoupling of feature extractor and classifier, and Per-FedAvg \cite{fallah2020personalized} a meta-learning approach specifically designed for solving PFL problems.

\subsection{Training setting.} 
To ensure fair and reliable comparisons, similar hyperparameters and model architectures have been used for all the baselines. 
For all the datasets, $80$ clients ($C=80$) are used for training the models and $20$ clients for evaluation of the performance. The number of communication rounds is set to $1000$, except for FEMNIST and Meta-Dataset where it is set to $2000$, due to the complexity of the datasets. For each round of communication, we randomly select $M=5$ clients, and we perform $S=5$ personalization steps. At test time, we let the models perform $50$ personalization steps to allow a good personalization to each client's own data. We use mini-batch stochastic gradient descent (SGD) as a local optimizer, with a fixed batch size of $30$ samples. 
For \emph{CAFeMe}, we use the feature-wise gating defined in Equation \ref{eqn:modulation} as a way to modulate the activations of the base model. However, we also investigated an alternative way by using an affine transformation, as used in FiLM \cite{perez2018film}. In this case, the modulation parameters $\zeta^l$ of a layer $l$ consists of two vectors, $\zeta_a^l$ and $\zeta_b^l$, used to scale and bias the activations, i.e., $$a(\mathbf{x} ~\psi^l) \odot \zeta^l = \zeta_a^l ~ \otimes ~ a(\mathbf{x} ~\psi^l) ~ + ~ \zeta_b^l.$$ Nonetheless, with this modulation, the performance of \emph{CAFeMe} is highly variable across different runs of the algorithm and doesn't achieve a good performance. %obtaining an average accuracy of $59.14 \pm 28.97$ instead of $90.66 \pm 0.66$ in CIFAR-10 with \emph{shards}, and $10.6 \pm 1.69$ instead of $74.97 \pm 1.67$ in CIFAR-10 with \emph{dirichlet} partition. 
This is likely because scaling the model's activations with a value that is not normalized in $[0,1]$ (unlike the sigmoidal feature-wise gating of Equation \ref{eqn:modulation}) may result in ending up in a local minima far from the optimal results. Therefore, in the rest of the experiments, all results are reported using the feature-wise gating mechanism in Equation \ref{eqn:modulation}.

We construct two different models for RMNIST/FEMNIST and CIFAR-10/Meta-Dataset, respectively. The first CNN model comprises two modules, each consisting of a convolutional layer with 32 filters, followed by batch normalization and ReLU nonlinearities. One linear layer with a size of 576 is used to complete the classification model. Similarly, the architecture of the federated modulator consists of the same two modules followed by three linear layers with sizes of 100, 200, 200, respectively. The second CNN model is similar to the first one but comprises three modules with 64 filters and two linear layers with a size of 576. Also, the federated modulator has three modules and four linear layers with sizes of 100, 100, 200, 200.

To account for statistical variations, we conduct five complete runs of the algorithms and report the results as the average accuracy and standard deviation.

\section{Results}

\begin{table*}[!htb]
\caption{Comparison of test accuracy (\%) after $50$ personalization steps on new clients at test time on synthetic datasets.}
\label{tab:synthetic_results}
\centering
\begin{tabular}{ lcccc} 
\hline
\textbf{Method} & \multicolumn{2}{c}{\textbf{CIFAR-10}} & \multicolumn{2}{c}{\textbf{RMNIST}} \\
\cline{2-3} \cline{4-5}
& \emph{Shards} & \emph{Dirichlet} & \emph{Shards} & \emph{Dirichlet} \\
\hline
FedAvg & $27.87 \pm 1.10$ & $40.91 \pm 1.98$ & $60.26 \pm 3.77$& $79.16 \pm 2.93$\\
FedAvg-FT & $87.00 \pm 2.40$ & $67.13 \pm 1.48$& $87.08 \pm 3.39$& $89.47 \pm 1.43$\\  
IFCA & $29.01 \pm 2.06$ & $42.10 \pm 3.00$& $65.13 \pm 6.26$& $80.50 \pm 1.69$\\
IFCA-FT & $87.26 \pm 2.50$ & $64.96 \pm 1.21$& $88.52 \pm 3.32$& $89.84 \pm 1.21$\\
Ditto & $89.87 \pm 1.02$& $70.71 \pm 1.54$& $90.44 \pm 1.44$& $89.27 \pm 1.50$\\
FedRep & $88.77 \pm 1.43$& $68.44 \pm 2.16$& $90.95 \pm 1.64$& $89.88 \pm 1.46$\\
Per-FedAvg & $88.41 \pm 2.06$& $71.04 \pm 1.91$& $83.86 \pm 1.74$& $74.57 \pm 3.12$\\
\textbf{CAFeMe} & $\mathbf{90.66 \pm 0.66}$& $\mathbf{74.97 \pm 1.67}$& $\mathbf{98.82 \pm 0.11}$& $\mathbf{94.31 \pm 1.00}$\\
\hline
\end{tabular}
\end{table*}

\begin{figure}[htb]
\centering
\includegraphics[width=0.48\textwidth]{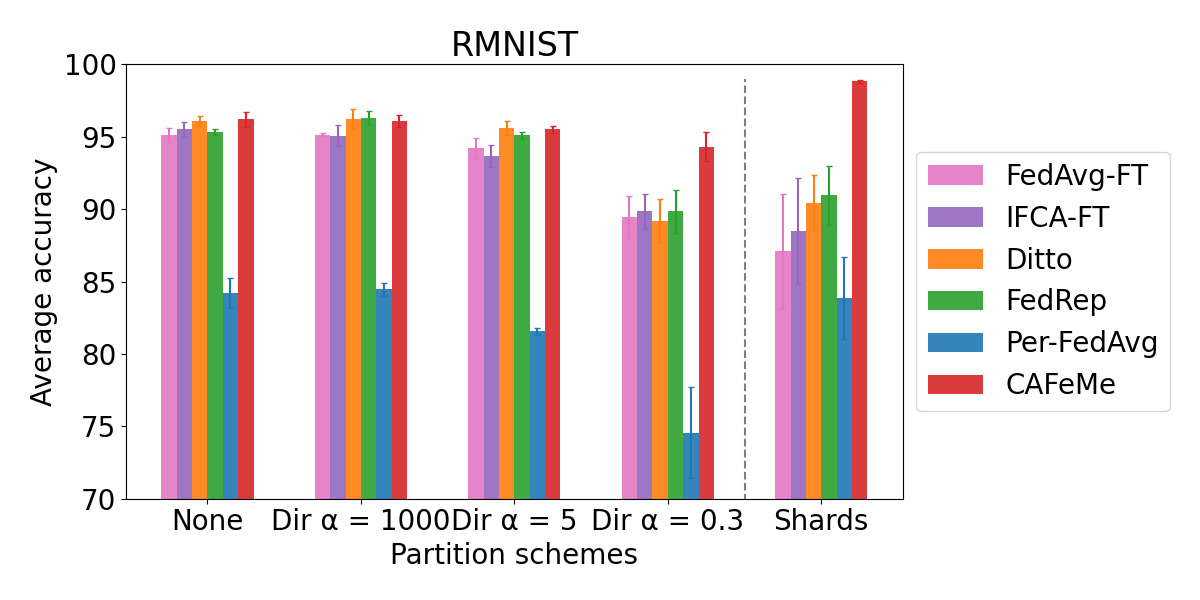}
\caption{Performance comparison on RMNIST dataset varying data heterogeneity. ``None" indicates that no partition schemes have been applied on the dataset, while ``Dir" stands for \emph{dirichlet} partition scheme with $\alpha$ the concentration parameter of the Dirichlet distribution.} \label{fig:data_hetero}
\end{figure}

\begin{table}[htb]
\caption{Comparison of test accuracy (\%) after $50$ personalization steps on new clients at test time on realistic datasets.}
\label{tab:realistic_results}
\centering
\begin{tabular}{ lccccc} 
\hline
\textbf{Method} & \textbf{FEMNIST} & \textbf{Meta-Dataset} \\
\hline
FedAvg & $79.81 \pm 2.21$ & $10.58 \pm 0.75$\\
FedAvg-FT & $82.07 \pm 1.92$& $57.78\pm 1.16$\\  
IFCA & $79.81 \pm 2.04$& $9.74 \pm 0.73$\\
IFCA-FT & $82.56 \pm 1.79$& $57.17 \pm 1.44$\\
Ditto & $80.41 \pm 1.29$& $45.31 \pm 1.44$\\
FedRep & $78.10 \pm 1.81$& $45.17 \pm 1.90$\\
Per-FedAvg & $80.26 \pm 2.60$& $57.89 \pm 1.09$\\
\textbf{CAFeMe} & $\mathbf{82. 76 \pm 0.83}$& $\mathbf{62.62 \pm 1.33}$\\
\hline
\end{tabular}
\end{table}
The experimental results are summarized in Tables \ref{tab:synthetic_results} and \ref{tab:realistic_results} for the synthetic and realistic datasets, respectively. Across all datasets, \emph{CAFeMe} consistently outperforms the other baseline methods. The performance gap becomes even more evident when the heterogeneity in clients' data distribution is large (as observed in RMNIST and Meta-Dataset), highlighting the effectiveness of the proposed approach in personalizing to each client's local data distribution. 
This observation is supported by the results presented in Figure \ref{fig:data_hetero}, which illustrates the performance of different PFL methods under varying levels of data heterogeneity in the RMNIST dataset.
\begin{figure}[htb]
\centering
\includegraphics[width=.45\textwidth]{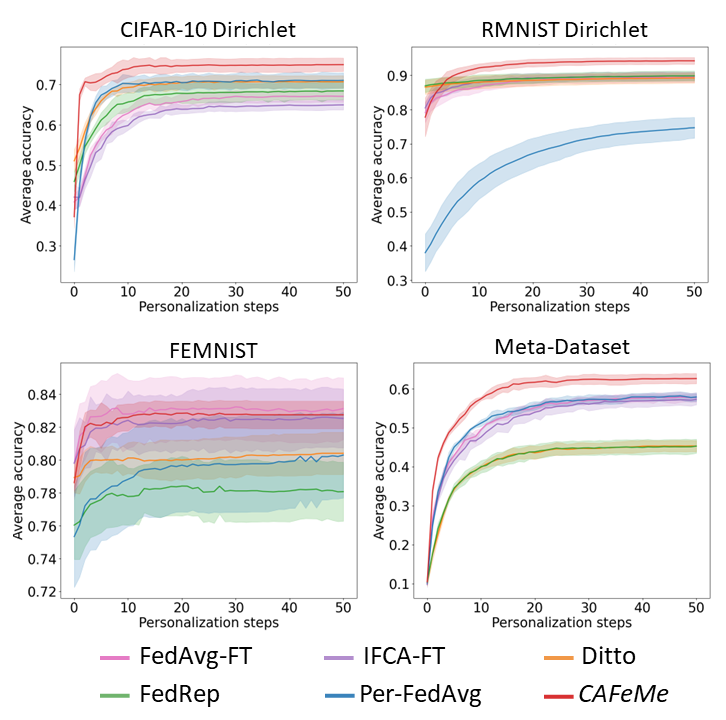}
\caption{Personalization to new clients at test time varying the number of personalization steps.}
\label{fig:pers_steps}
\end{figure}
When the data are (almost) balanced and i.i.d, as indicated by the ``None" column in the histogram and when $\alpha=1000$, the performance of \emph{CAFeMe} is comparable to other baselines, such as Ditto and FedRep. However, as the data heterogeneity across clients increases (lower $\alpha$ values), \emph{CAFeMe} consistently maintains strong performance, while the performance of the other baselines significantly deteriorates. This is particularly evident when using the \emph{shards} partition scheme, where most of the clients have data from only two classes. In this challenging scenario, personalizing the global model to each local dataset becomes difficult due to the narrow local data distribution compared to the one used for learning the global model. Nevertheless, \emph{CAFeMe} overcomes this challenge by effectively modulating the global model through meta-learning, resulting in robust personalization even in the presence of concept shifts. This observation is further supported by the results on Meta-Dataset (Table \ref{tab:realistic_results}). The variability in the conditional distribution $P_{\mathcal{Y}|\mathcal{X}}$ across different clients hurts the performance of other methods, which struggle to adapt the global model to such variability. Even using multiple global models, as in IFCA-FT, is insufficient for effective personalization on the client side, also due to the limitations on the estimation of the cluster identity. In contrast, \emph{CAFeMe} leverages meta-learning and the federated modulator to achieve superior performance in personalizing the global model among clients with concept shifts. 

\begin{figure}[tbp]
\centering
\includegraphics[width=0.45\textwidth]{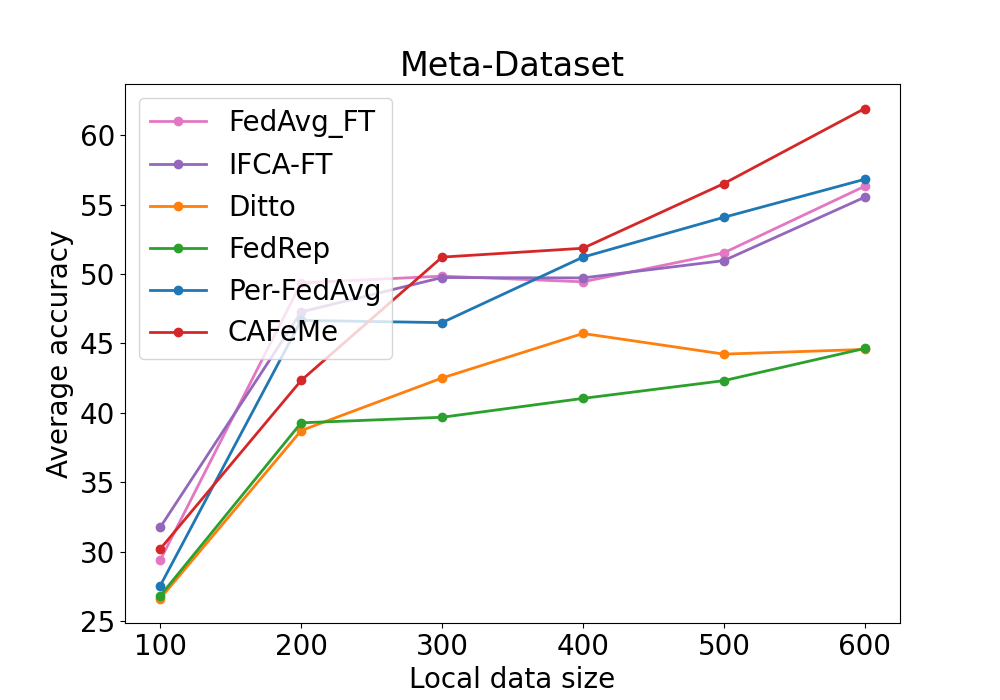}
\caption{Performance comparison on Meta-Dataset varying the size of data available at the client level. } \label{fig:varying_size}
\end{figure}

Furthermore, Figure \ref{fig:pers_steps} shows that \emph{CAFeMe} achieves efficient personalization with a small number of personalization steps at test time across all datasets. This demonstrates the efficiency of the proposed approach, enabling quick personalization for clients with limited computational resources. We also conducted additional experiments to investigate the impact of different dataset sizes at the client level. As depicted in Figure \ref{fig:varying_size}, when the local data size is too small (e.g., 100), all baselines exhibit low performance due to the inability of personalizing the global model effectively to the client's local data distribution. On the other hand, as the size of data increases, \emph{CAFeMe} shows a substantial improvement in performance, suggesting its applicability even in scenarios with small amounts of data at the client level.

\section{Conclusion}
In this work, we presented \emph{CAFeMe}, a novel approach for PFL that leverages meta-learning and a federated modulator network to achieve efficient and effective personalization of the global model to each client's local data distribution. Our proposed approach overcomes the limitations of traditional FL methods in scenarios with non-i.i.d data distributions by introducing modulation layers that condition the base model on each client's data.

Through extensive experiments on various synthetic and realistic datasets, we demonstrated the superiority of \emph{CAFeMe} over state-of-the-art baselines. The performance gap was particularly evident when the data heterogeneity among clients was significant, showing the robustness and adaptability of our approach in such challenging scenarios. The ability to personalize the global model effectively, even in the presence of concept shifts, was one of the key strengths of \emph{CAFeMe}, setting it apart from traditional approaches that struggled to handle such scenarios. We believe that \emph{CAFeMe} holds great promise for advancing personalized federated learning in various real-world applications, and we hope that our work inspires further research in this exciting and rapidly evolving field.
Future research directions may involve investigating more sophisticated modulator architectures as well as diverse meta-learning techniques and algorithms to optimize the personalization process for clients with varying data distributions. Furthermore, it would be interesting to explore the robustness of \emph{CAFeMe} under adversarial settings and develop mechanisms to defend against potential security and privacy threats. 

\section*{Acknowledgments}
This work was supported by the ``Knowledge Foundation'' (KK-stiftelsen).

\printbibliography

\clearpage

\begin{multicols}{3}
[
\title{\Large \textbf{Supplementary Material}}
]
\end{multicols}
\section{Datasets}

\subsection{Synthetic datasets.}
We evaluate our model on two synthetic datasets: CIFAR-10 and a rotated version of MNIST, referred to as RMNIST. For the latter, we partition MNIST data into $10$ subsets of equal size and we apply a distinct degree of rotation within the range $[0, 200]$ to each subset. This rotational variation enhances the diversity within the MNIST dataset, enabling a more realistic representation of non-i.i.d. data. 

\begin{figure}[!ht]
\centering
\subfloat[][Shards]
{\includegraphics[width=.35\textwidth]{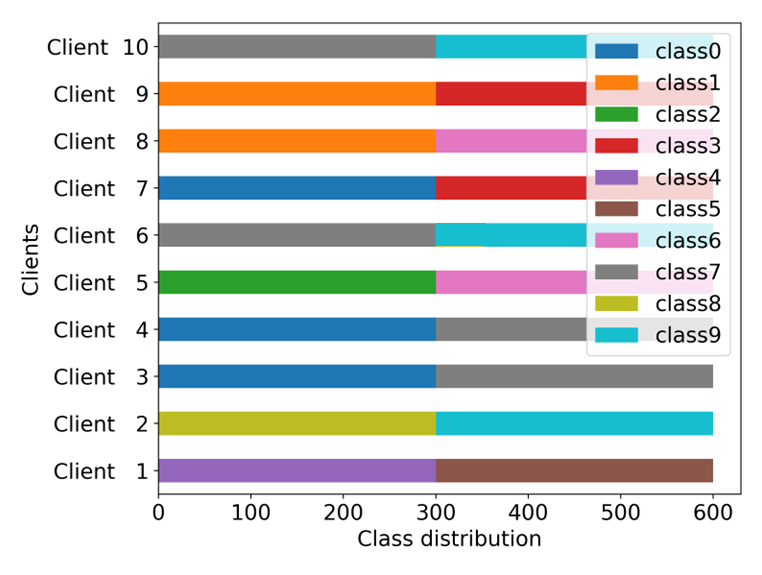}} \\
\subfloat[][Dirichlet $\alpha=1000$]
{\includegraphics[width=.35\textwidth]{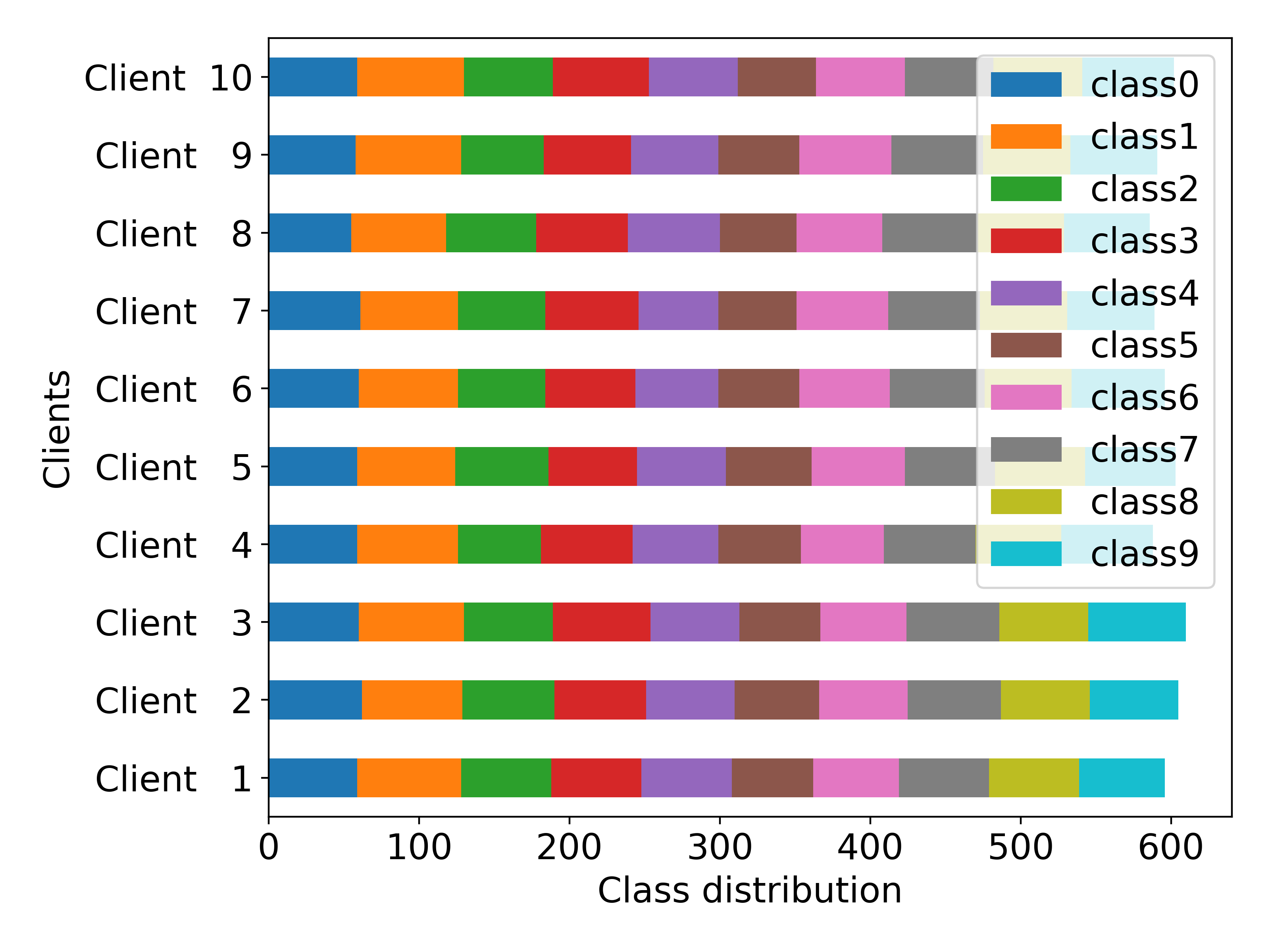}} \\
\subfloat[][Dirichlet $\alpha=0.3$]
{\includegraphics[width=.35\textwidth]{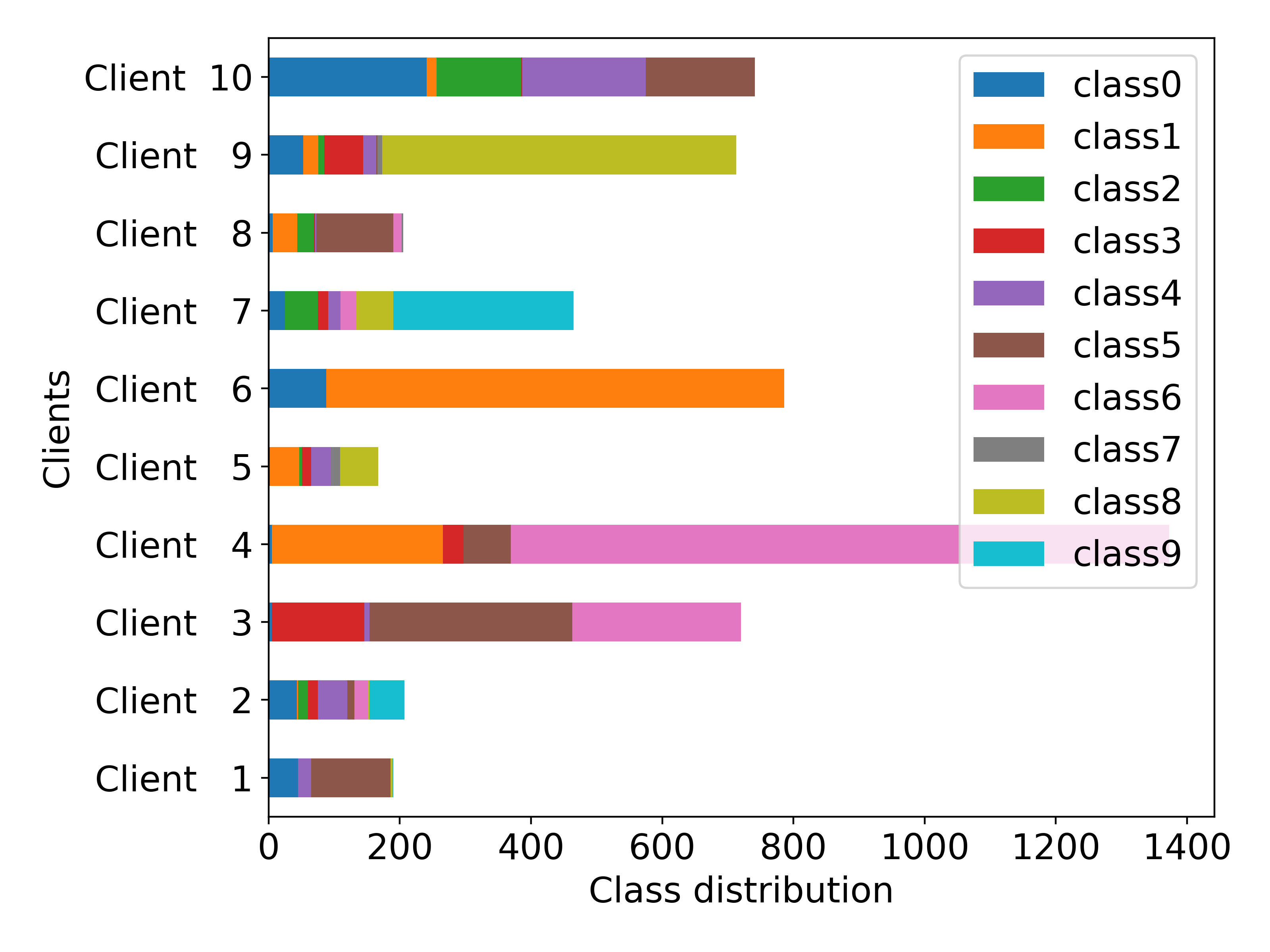}} 
\caption{Class distributions of 10 exemplary clients using different partition schemes for RMNIST dataset.}
\label{fig:partition}
\end{figure}

To partition data into clients, we designed two partition schemes: the \emph{shards} partition and the \emph{dirichlet} partition. A visual representation of these two partition schemes is shown in Figure \ref{fig:partition}. 
The \emph{shards} partition scheme involves a balanced non-i.i.d. partition, where each client is allocated an equal number of samples. To construct this scheme, we first sort the data by label, divide it into $200$ shards, and assign two shards to each client. Therefore, most clients solely have examples from two classes. On the other hand, the \emph{dirichlet} partition exhibits an unbalanced and non-i.i.d. scheme, characterized by varying numbers of data points and class proportions across clients. For each client, we sample $\mathbf{p}_n \sim \textnormal{Dir}_C(\alpha)$, where $C$ represents the number of clients, and allocate a $\mathbf{p}_{n,i}$ proportion of class $n$ to local client $i$. The concentration parameter $\alpha$ regulates the level of data heterogeneity across different clients: the smaller this value, the higher the class imbalance. We set it to $0.3$ by default if not explicitly mentioned. 

\subsection{Realistic datasets.}
To assess the performance of \emph{CAFeMe} in a more realistic scenario we evaluate it on FEMNIST and Meta-Dataset. While FEMNIST is a well-known dataset in the field of PFL, this is the first work that uses Meta-Dataset to assess the performance of a PFL framework. 
Meta-Dataset is a large-scale dataset comprising various classification datasets such as Mini-Imagenet \cite{ravi2017optimization}, FC100 \cite{oreshkin2018tadam}, Omniglot \cite{lake2011one}, Aircraft \cite{maji2013fine}, CUB Birds \cite{wah2011caltech}, Describable Textures Dataset (DTD) \cite{cimpoi2014describing}, Traffic Signs \cite{stallkamp2012man}, and VGG Flowers \cite{nilsback2008automated}. In this context, each client is assigned data with $N$ classes sampled from one of these classification datasets. Therefore, while all clients share the same task of classifying images into $N$ classes, the nature of their local data varies significantly, resulting in labels that might have different meanings across clients. This leads to the emergence of the concept shift scenario, a common phenomenon in real-world situations.

\end{document}